\documentclass{article} % For LaTeX2e
\usepackage{colm2024_conference}

\usepackage{amsmath,amsfonts,bm}

\def\eqref#1{equation~\ref{#1}}
\def\1{\bm{1}}

\DeclareMathAlphabet{\mathsfit}{\encodingdefault}{\sfdefault}{m}{sl}
\SetMathAlphabet{\mathsfit}{bold}{\encodingdefault}{\sfdefault}{bx}{n}

\usepackage{hyperref}
\usepackage{xurl}
\usepackage{booktabs}
\usepackage{graphicx}
\usepackage{multirow}
\usepackage{colortbl}
\usepackage{xcolor}
\usepackage{algorithm}
\usepackage{algorithmic}
\usepackage{amsmath}
\usepackage{amssymb}
\usepackage{xspace}
\usepackage{caption}
\usepackage{wrapfig}
\usepackage{float}
\usepackage{placeins}
\usepackage{needspace}
\usepackage{fontawesome5}
\definecolor{githubblue}{RGB}{0,0,128}
\renewenvironment{abstract}{\par\vspace{2pt}\centerline{\large\bfseries Abstract}\vspace{4pt}\noindent\ignorespaces}{\par\vspace{4pt}}
\usepackage{enumitem}
\usepackage{tocloft}

\cftsetindents{section}{0em}{1.5em}
\cftsetindents{subsection}{1.5em}{2.3em}
\cftsetindents{subsubsection}{3.8em}{3.2em}

\title{MC-Sparse: Deconstructing and Closing \\ the Dense--Sparse Attention Gap in Diffusion Transformers}

\author{
\vspace{-1.0em}
{\href{https://dodododddo.github.io/mcsparse-project-page/}{\resizebox{0.62\textwidth}{!}{\textcolor{black}{\faGithub}\hspace{0.5em}\textcolor{githubblue}{\texttt{https://dodododddo.github.io/mcsparse-project-page/}}}}}\\[0.85em]
Jiarui Chen$^{1,2,3}$, Zeqiang Lai$^{2,4\dagger}$,  Jiangshan Wang$^{2,4}$,\\
Ziheng Ouyang$^{2,5}$, Ye Huang$^{6}$, Xiangyu Yue$^{4}$,\\ Cewu Lu$^{3,7*}$, Chunchao Guo$^{2*}$ \\[0.5em]
$^{1}$Fudan University\quad $^{2}$Tencent HY \quad
$^{3}$Shanghai Innovation Institute \\
$^{4}$MMLab, CUHK \quad $^{5}$Nankai University \\
$^{6}$Peking University \quad $^{7}$Shanghai Jiao Tong University \\
}

\begin{document}

\maketitle

% Author-role symbols follow Tex-Zero_arxiv: dagger = project lead;
% asterisk = corresponding authors. Keep the note local to the title page.
\begingroup
\renewcommand{\thefootnote}{}
\footnotetext[0]{\textsuperscript{$\dagger$} Project lead. \textsuperscript{$*$} Corresponding authors.}
\endgroup

\begin{abstract}
Sparse attention is a primary approach to reducing the latency of diffusion transformers in long-sequence generation tasks, such as video and high-resolution 3D asset generation. However, existing methods can degrade generation quality and fidelity at high sparsity levels. Through controlled oracle comparisons, we trace this degradation to three sources: constraints imposed by token grouping, inaccurate interaction selection, and the attention contributions lost when tokens are discarded. Guided by this analysis, we propose Meta-Cached Sparse Attention (MC-Sparse), a training-free framework that selects individual key-value (KV) tokens while organizing similar queries into tile-aligned groups for efficient GPU execution. MC-Sparse caches metadata comprising query groups, KV indices selected using exact attention probabilities, and residuals between dense and sparse attention outputs, and reuses them across subsequent denoising steps. Across video and 3D generation models, MC-Sparse achieves higher fidelity to dense-attention outputs and larger denoising speedups than existing sparse-attention baselines, without visible quality degradation. Relative to dense attention, it delivers a $1.80\times$ denoising speedup on Minimax-H3-Base and a $2.32\times$ speedup on 3D asset generation, both with negligible quality loss.
\end{abstract}

\begin{figure}[H]
    \captionsetup{font=small,skip=3pt}
    \centering
    \includegraphics[width=\linewidth]{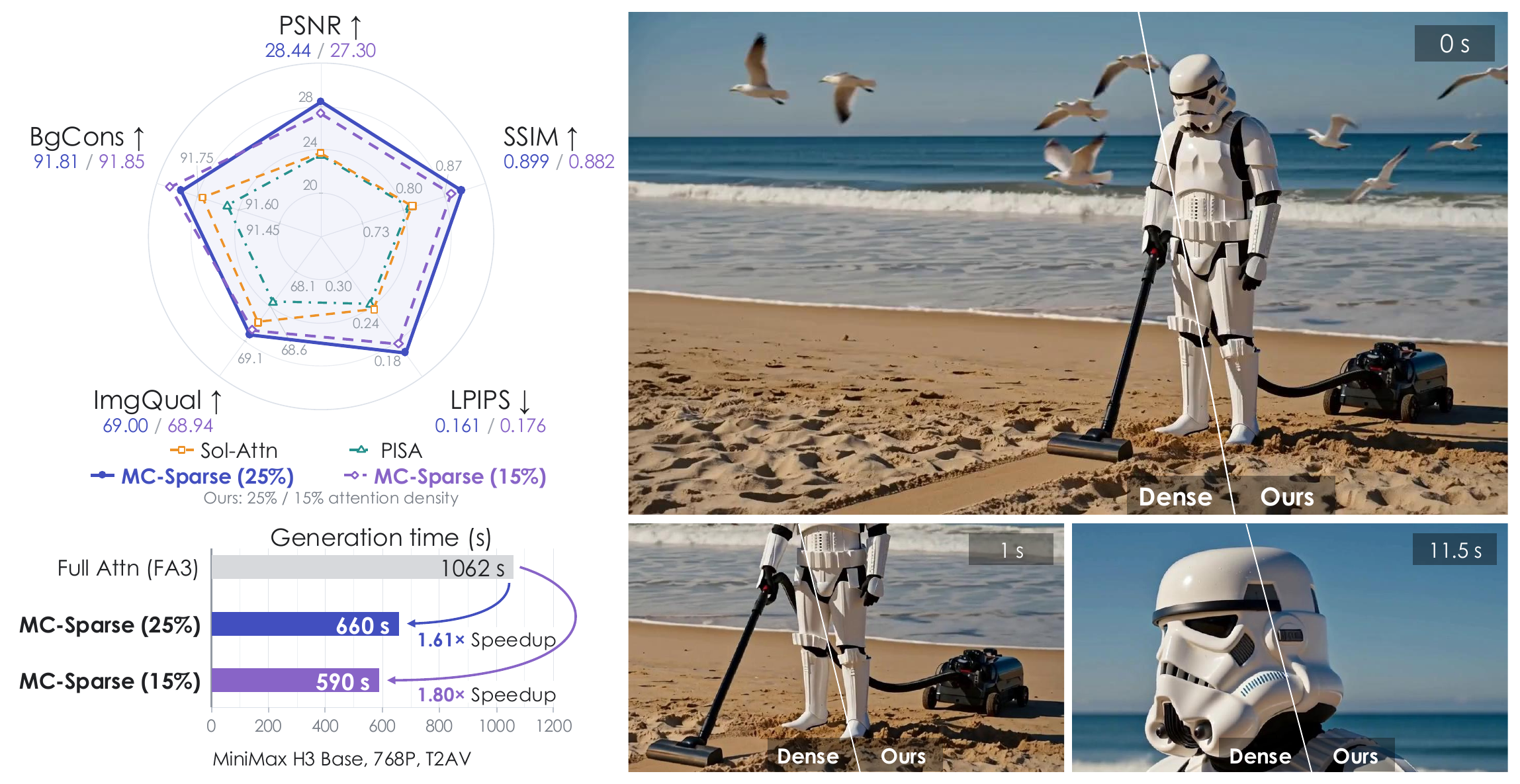}
    \caption{Quality and efficiency of MC-Sparse on Minimax-H3-Base at 768p.
Left: quality metrics and DiT denoising latency. At 15\% and 25\% attention density, MC-Sparse achieves $1.80\times$ and $1.61\times$ speedup over dense attention (FA3), respectively. 
Right: frames generated at 15\% density closely match the dense-attention reference across the video.
}
    \label{fig:teaser}
\end{figure}

% \clearpage
% \tableofcontents
% \clearpage

% Required for the itemize options below: \usepackage{enumitem}.
\section{Introduction}

% Diffusion transformers~\citep{peebles2023scalablediffusionmodelstransformers}
% have become a dominant backbone for high-fidelity video and 3D asset
% generation. Their increasing sequence lengths make attention a major inference
% bottleneck: the quadratic cost of processing spatiotemporal video tokens and
% geometric 3D tokens grows rapidly with output resolution and video duration.
% Sparse attention offers a natural way to reduce this cost, as attention often
% concentrates on a small subset of token interactions. Yet aggressive sparsity
% frequently degrades generation quality, leaving a persistent gap between
% sparse-attention efficiency and dense-attention fidelity.

Diffusion transformers~\citep{peebles2023scalablediffusionmodelstransformers} have become a dominant backbone for high-fidelity video and 3D asset generation\citep{wan2025wanopenadvancedlargescale, minimax2026h3,lai2025latticedemocratizehighfidelity3d, lai2025hunyuan3d25highfidelity3d}. However, their growing sequence lengths make attention a major inference bottleneck: its quadratic cost scales rapidly with video duration and output resolution, since it must process spatiotemporal video tokens and geometric 3D tokens. Sparse attention offers a natural remedy, as attention often concentrates on a small subset of token interactions. Yet aggressive sparsity frequently degrades generation quality, leaving a persistent gap between the efficiency of sparse attention and the fidelity of dense attention.

% Most existing methods impose sparsity at block
% granularity~\citep{zhang2025spargeattentionaccuratetrainingfreesparse,li2026pisapiecewisesparseattention,li2026solattnacceleratingvideogeneration}.
% Block-sparse attention (BSA) selects entire KV blocks for groups of queries,
% allowing sparse computation to use efficient GPU tiles. Practical selectors
% often estimate block importance from pooled query and key representations.
% This couples approximate scoring with a constrained selection layout, making
% it difficult to determine how much of the quality gap comes from the selector,
% how much comes from the block structure, and how much remains even when both
% are improved.

Most existing methods impose sparsity at block granularity~\citep{xi2025sparsevideogenacceleratingvideo, zhang2025vsafastervideodiffusion}. Block-sparse attention (BSA) selects entire KV blocks for each group of queries, which maps well onto efficient GPU tiles. In practice, block importance is typically estimated from pooled query and key representations\citep{li2026pisapiecewisesparseattention, li2026solattnacceleratingvideogeneration}. This design entangles two sources of error: the approximate scores used to pick blocks, and the coarse block layout that restricts which interactions can be picked at all. As a result, it is hard to tell how much of the quality gap stems from the selector, how much from the block structure, and how much persists even when both are improved.

% We examine these questions through controlled oracle comparisons at matched
% attention densities. Each oracle maximizes retained attention mass using
% exact dense-attention probabilities: the block oracle selects KV blocks,
% the token oracle selects individual KV tokens for shared query groups, and
% the per-query oracle additionally allows each query its own selection.
% Comparisons with mean-pooled BSA and among these oracles reveal three sources
% of the gap: \emph{structural binding error} from coupled KV tokens and shared
% query selections, \emph{selection error} from approximate scoring, and
% \emph{discarded tail error} from nonzero contributions of omitted tokens.
% These findings motivate finer-grained selection, accurate attention scoring,
% and compensation for discarded contributions.

\begin{figure}[t]
    \centering
    \includegraphics[width=\linewidth]{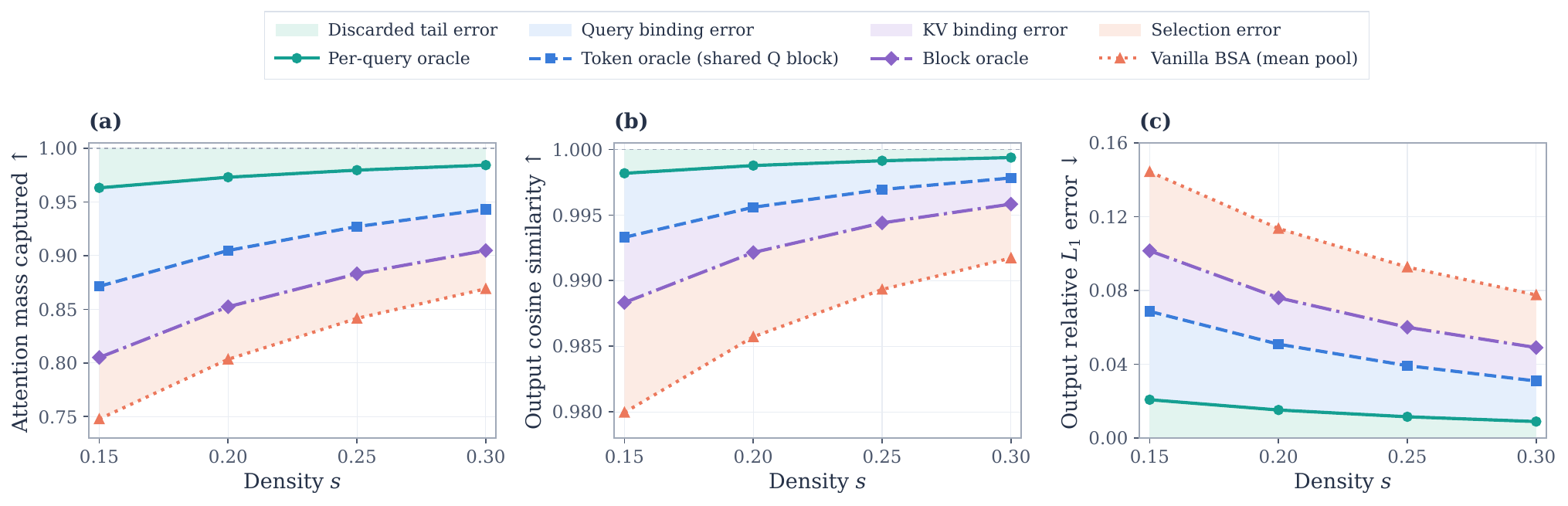}
    \caption{Oracle comparisons of the dense-to-sparse gap on Wan2.1-1.3B-T2V at 480p. Experimental settings are provided in Appendix~\ref{app:experimental_details}.}
    \label{fig:oracle_combined}
\end{figure}

We examine these questions through controlled oracle comparisons at matched attention densities (i.e., the same sparsity level). Each oracle maximizes the retained attention mass using exact dense-attention probabilities, but differs in selection granularity: the \emph{block oracle} selects oracle KV blocks for each query group; the \emph{token oracle} selects or oracle individual KV tokens for each query group; and the \emph{per-query oracle} further lets every query choose its own KV tokens. Comparing mean-pooled BSA with these oracles, and the oracles with one another, isolates three sources of the quality gap (Fig.~\ref{fig:oracle_combined}):\emph{structural binding error}, which arises because KV tokens are bound together in blocks and different queries in the same group share a single selection; \emph{selection error}, which arises from approximate scoring (e.g. mean pooling); and \emph{discarded tail error}, which arises because omitted tokens still carry nonzero attention mass. These findings motivate three design principles: finer-grained selection, accurate attention scoring, and compensation for discarded contributions.

% Implementing these designs efficiently poses several challenges. Exact
% token-level selection requires expensive attention scoring, while gathering
% individual KV tokens introduces irregular memory access. Queries must still
% be grouped into sufficiently large tiles for efficient GPU execution.
% Unconstrained clustering groups similar queries to reduce binding, but
% produces variable-size groups that incur redundant computation from padding
% inside the attention kernel to match its computation
% tiles~\citep{zhou2026svgearparameterfreelinearcompensation}.
% Token-level selection also removes the natural block structure used to
% summarize discarded regions for compensation.

Realizing these principles efficiently is challenging. First, exact token-level selection requires computing attention scores, which is expensive, and gathering individual KV tokens causes irregular memory access. Second, queries must still be grouped into sufficiently large tiles for efficient GPU execution. Unconstrained clustering can group similar queries to reduce structural binding, but it yields variable-size groups that must be padded to fill the kernel's computation tiles, wasting computation~\citep{zhou2026svgearparameterfreelinearcompensation}. Third, token-level selection removes the block structure that would otherwise provide a natural way to summarize discarded regions for compensation.

% Temporal redundancy offers a way to amortize accurate selection and support
% compensation. We find that exact selections from earlier denoising steps can
% outperform mean-pooled selection recomputed at every step, while the
% dense--sparse output residual exhibits small absolute changes across nearby
% steps. Building on these observations, we propose \textbf{Meta-Cached Sparse
% Attention (MC-Sparse)}, a training-free framework that combines token-level
% KV selection, tile-aligned query grouping, and cross-step reuse. At anchor
% steps, MC-Sparse groups similar queries into equal-size tiles, selects KV
% tokens using exact group-wise attention mass, and records the dense--sparse
% output residual. Subsequent steps reuse the query groups, selected indices,
% and residuals while recomputing the current queries, keys, and values. This
% amortizes grouping and exact selection and uses cached residuals for temporal
% tail compensation. Fast PDDP, a two-pass exact selector, and a token-sparse
% attention kernel support efficient grouping, selection, and execution.

Temporal redundancy across denoising steps offers a way to amortize exact selection and to support compensation. We find that reusing exact selections from earlier steps can outperform recomputing mean-pooled selections at every step, and that the dense--sparse output residual changes little between nearby steps. Building on these observations, we propose \textbf{Meta-Cached Sparse Attention (MC-Sparse)}, a training-free framework that combines token-level KV selection, tile-aligned query grouping, and cross-step reuse. At anchor steps, MC-Sparse groups similar queries into equal-size tiles, selects KV tokens according to their exact attention mass aggregated over each query group, and records the dense--sparse output residual. Subsequent steps reuse the query groups, selected indices, and cached residuals, while computing attention with the current queries, keys, and values. This amortizes the cost of grouping and exact selection, and the cached residuals compensate for the discarded tail across steps. To make this efficient in practice, we develop fast Principal
Direction Divisive Partitioning(PDDP)\citep{boley98principal} for grouping, a two-pass exact selector for selection, and a token-sparse attention kernel for execution.

% Across the evaluated video models, MC-Sparse achieves higher fidelity to
% dense-attention outputs and greater measured denoising speedups than the
% compared sparse-attention baselines while preserving visual quality.
% On Minimax-H3-Base, MC-Sparse at 15\% attention density reaches $1.80\times$
% DiT denoising speedup over dense attention while retaining higher fidelity
% than competing sparse methods. On HY3D-Internal, MC-Sparse achieves $2.32\times$ denoising
% speedup at 15\% attention density with negligible quality degradation.

Across the evaluated video models, MC-Sparse achieves higher fidelity to dense-attention outputs and larger denoising speedups than the compared sparse-attention baselines, while preserving visual quality. On Minimax-H3-Base, at 15\% attention density, it reaches a $1.80\times$ DiT denoising speedup over dense attention with higher fidelity than competing sparse methods. On HY3D-Internal, at the same density, it achieves a $2.32\times$ denoising speedup with negligible quality degradation.

% Our contributions are summarized as follows:
% \begin{itemize}[leftmargin=1em, noitemsep, topsep=0pt, parsep=0pt]
%     \item We use controlled oracle comparisons to distinguish structural
%     binding error, selection error, and discarded tail error, quantifying
%     their effects on attention fidelity at matched budgets.
%     \item We propose MC-Sparse, combining token-level KV selection and
%     tile-aligned query grouping with cross-step reuse of exact selections
%     and output residuals, supported by efficient GPU implementations.
%     \item We demonstrate the effectiveness and generality of MC-Sparse
%     across multiple diffusion transformer models and tasks spanning video
%     and 3D asset generation.
% \end{itemize}

Our contributions are summarized as follows:
\begin{itemize}[leftmargin=1em, noitemsep, topsep=0pt, parsep=0pt]
    \item We use controlled oracle comparisons to disentangle three sources of quality degradation in sparse attention: structural binding error, selection error, and discarded tail error, and quantify how each affects attention fidelity at matched attention densities.
    \item We propose MC-Sparse, a training-free framework that combines token-level KV selection and tile-aligned query grouping with cross-step reuse of exact selections and output residuals, together with efficient GPU implementations.
    \item We demonstrate the effectiveness and generality of MC-Sparse on multiple diffusion transformer models, spanning both video and 3D asset generation.
\end{itemize}
\section{Related Work}

\subsection{Long-Sequence Diffusion Transformers}

Diffusion transformers (DiTs)~\citep{peebles2023scalablediffusionmodelstransformers} are widely used in content
generation, with video and 3D asset generation representing two
prominent long-sequence tasks. Representative models include
HunyuanVideo~\citep{kong2025hunyuanvideosystematicframeworklarge}, Wan2.1~\citep{wan2025wanopenadvancedlargescale}, and
MiniMax-H3~\citep{minimax2026h3} for video generation, alongside recent advances in 3D asset generation~\citep{lai2025hunyuan3d25highfidelity3d,
lai2025latticedemocratizehighfidelity3d,
xiang2025nativecompactstructuredlatents,
wang2026doesnative3dtexture,wang2026dreamstyle3d}. As generation resolution
increases, these tasks require longer token sequences, which can reach
tens to hundreds of thousands of tokens,
making the quadratic cost of full attention a dominant contributor to
inference latency.

\subsection{Efficient Attention}

\textbf{Trainable efficient attention.}
Trainable approaches include linear attention methods such as
SANA~\citep{xie2024sanaefficienthighresolutionimage}, SANA-Video~\citep{chen2025sanavideoefficientvideogeneration}, and
MHLA~\citep{zhang2026mhlarestoringexpressivitylinear}, and sparse attention methods such as
VMoBA~\citep{wu2025vmobamixtureofblockattentionvideo}, VSA~\citep{zhang2025vsafastervideodiffusion}, and
SpargeAttention2~\citep{zhang2026spargeattention2trainablesparseattention}, as well as hybrid attention
methods such as SLA~\citep{zhang2025slasparsitydiffusiontransformers} and SLA2~\citep{zhang2026sla2sparselinearattentionlearnable}.

\textbf{Training-free sparse attention.}
Training-free sparse attention instead aims to approximate full attention in
pretrained models while evaluating fewer token interactions. Many approaches
organize computation into blocks to exploit FlashAttention-style
tiling~\citep{dao2022flashattentionfastmemoryefficientexact}.
Block selection uses structured patterns or content-dependent importance
estimates, as explored by SVG~\citep{xi2025sparsevideogenacceleratingvideo},
XAttention~\citep{xu2025xattentionblocksparseattention}, and
SpargeAttention~\citep{zhang2025spargeattentionaccuratetrainingfreesparse}.
SVG-EAR~\citep{zhou2026svgearparameterfreelinearcompensation} and PISA~\citep{li2026pisapiecewisesparseattention} further
approximate contributions from unselected blocks.
Block construction commonly forms tile-aligned groups based on spatial or
spatiotemporal locality, or after serialization along Hilbert or Z-order
curves, with SpargeAttention using Hilbert ordering; semantic clustering in
SVG2~\citep{yang2026sparsevideogen2acceleratevideo}, SVOO~\citep{luo2026attentionsparsityinputstabletrainingfree}, and
AdaCluster~\citep{tan2026adaclusteradaptivequerykeyclustering} instead
produces variable-sized groups whose mismatch with kernel tile boundaries
introduces wasted computation within partially filled tiles, reducing actual
execution efficiency even without explicit padding.
A complementary line of work exploits cross-step redundancy in attention,
a strategy also used in diffusion caching~\citep{wang2026elasticdiffusiontransformer, wang2026precisecacheprecisefeaturecaching}.
AdaSpa~\citep{xia2025trainingfreeadaptivesparseattention},
DiTFastAttn~\citep{yuan2024ditfastattnattentioncompressiondiffusion}, and
QuantSparse~\citep{feng2026quantsparsecomprehensivelycompressingvideo}
reuse attention-related information along the denoising trajectory.

\begin{figure}[t]
    \centering
    % \captionsetup{skip=3pt}
    \includegraphics[width=\linewidth]{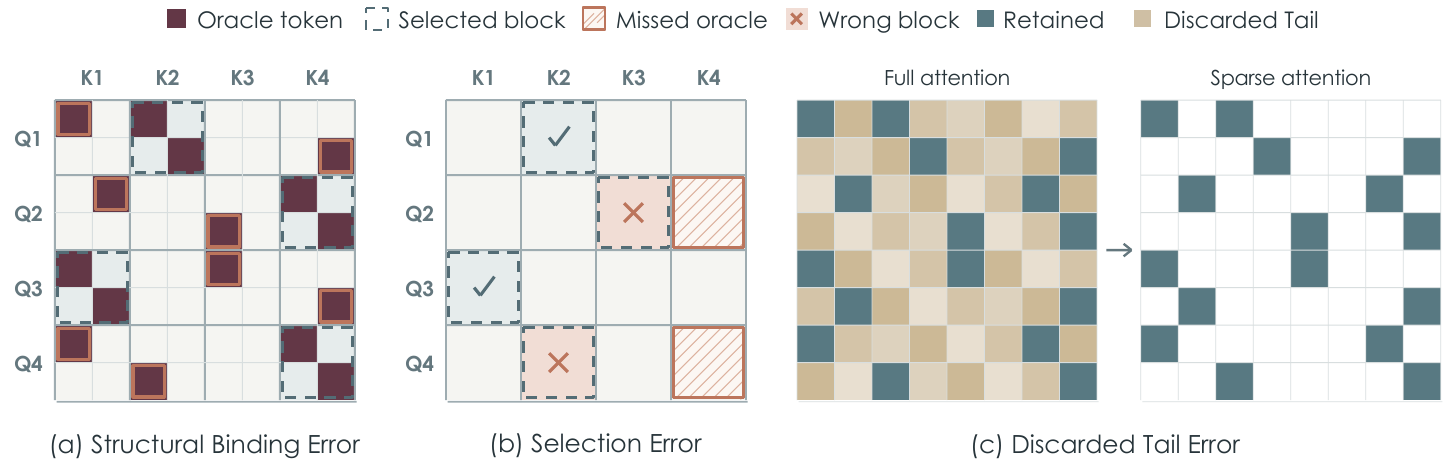}
    \caption{Three sources of the gap between dense and sparse attention.}
    \label{fig:gap_sources}
\end{figure}
\section{Preliminaries}
% \textbf{Dense attention.}
% Diffusion transformers typically use multi-head attention. For a single head
% with query, key, and value matrices $\mathbf Q$, $\mathbf K$, and $\mathbf V$,
% and query/key dimension $d$, dense attention computes
% $\mathbf O^{\mathrm{full}}=\operatorname{Softmax}(\mathbf Q\mathbf K^\top/\sqrt d)\mathbf V$.
% Softmax operates over keys, and $A_{ij}$ denotes the resulting dense attention
% weight from query $i$ to key $j$.

% \textbf{Block-sparse attention.}
% Sparse attention computes
% $\mathbf O^{\mathrm{sparse}}=\operatorname{Softmax}(\mathbf Q\mathbf K^\top/\sqrt d+\mathbf M)\mathbf V$,
% where $M_{ij}=0$ retains a query--key interaction and $M_{ij}=-\infty$ masks it
% out. For efficient GPU execution, block-sparse attention (BSA) retains entire
% KV blocks and shares one selection among queries in each block, with block
% sizes typically aligned with GPU computation tiles.

% \textbf{Vanilla BSA.} 
% We refer to BSA using a mean-pooled selector as \emph{vanilla BSA}.
% At each diffusion step, this selector scores block pairs by
% $\hat z_{gr}=\bar{\mathbf q}_g^\top\bar{\mathbf k}_r/\sqrt d$, where
% $\bar{\mathbf q}_g$ and $\bar{\mathbf k}_r$ are block means, and retains the
% highest-scoring KV blocks per query block under a given budget.

\textbf{Dense attention.}
Diffusion transformers typically use multi-head attention. For a single head with query, key, and value matrices $\mathbf Q$, $\mathbf K$, and $\mathbf V$, and query/key dimension $d$, dense attention computes
$\mathbf O^{\mathrm{full}}=\operatorname{Softmax}(\mathbf Q\mathbf K^\top/\sqrt d)\mathbf V$,
where the softmax is taken over keys. We write $A_{ij}$ for the resulting attention weight from query $i$ to key $j$.

\textbf{Block-sparse attention.}
Sparse attention restricts each query to a subset of keys:
$\mathbf O^{\mathrm{sparse}}=\operatorname{Softmax}(\mathbf Q\mathbf K^\top/\sqrt d+\mathbf M)\mathbf V$,
where $M_{ij}=0$ retains the query--key interaction and $M_{ij}=-\infty$ masks it out. For efficient GPU execution, block-sparse attention (BSA) partitions queries and keys into blocks, retains entire KV blocks, and shares one selection among all queries in a block. The block size is typically aligned with the GPU computation tile.

\textbf{Vanilla BSA.}
We refer to BSA with a mean-pooled selector as \emph{vanilla BSA}. At each diffusion step, this selector scores each query-block/KV-block pair $(g, r)$ by
$\hat z_{gr}=\bar{\mathbf q}_g^\top\bar{\mathbf k}_r/\sqrt d$,
where $\bar{\mathbf q}_g$ and $\bar{\mathbf k}_r$ are the mean-pooled query and key of blocks $g$ and $r$, respectively. For each query block, it retains the highest-scoring KV blocks under a given attention density.

\section{Method}

% We begin by examining the gap between dense attention and vanilla BSA.
% An \emph{oracle selector} maximizes retained attention mass using exact
% dense-attention probabilities under a fixed budget and grouping constraints.
% In Fig.~\ref{fig:oracle_combined}, the \emph{block oracle} selects KV blocks
% for each query block; the \emph{token oracle (shared Q block)} selects
% individual KV tokens but retains a shared selection within each query block;
% the \emph{per-query oracle} selects KV tokens independently for each query.
% At matched densities, these comparisons distinguish structural binding error,
% selection error, and discarded tail error. We analyze each source and introduce
% a corresponding design, then present the overall MC-Sparse pipeline and key
% implementation details in Section~\ref{sec:pipeline}

We first examine where vanilla BSA loses fidelity relative to dense attention. An \emph{oracle selector} retains the entries with the largest dense-attention probability mass under a fixed attention density. We consider three oracles (Fig.~\ref{fig:oracle_combined}): the \emph{block oracle} selects KV blocks for each query block; the \emph{token oracle (shared Q block)} selects individual KV tokens but shares one selection within each query block; and the \emph{per-query oracle} selects KV tokens independently for each query. At matched densities, comparing vanilla BSA with these oracles and with dense attention isolates three sources of the gap: structural binding error, selection error, and discarded tail error.

Because exact selection and query grouping are costly, MC-Sparse exploits the stability of attention across denoising steps: it performs them only at \emph{anchor steps} and reuses the resulting query groups, KV indices, and dense--sparse output residual at subsequent \emph{reuse steps}. In the following subsections, we analyze each error in turn and introduce a corresponding design.

\subsection{Mitigating Structural Binding Error}

% \textbf{Structural binding error.}
% \emph{KV binding} retains irrelevant neighbors of important tokens, while
% \emph{query binding} forces queries with different attention patterns to share
% one selection (Fig.~\ref{fig:gap_sources}(a)). The successive gains from block
% oracle to token oracle with shared query blocks, and then to per-query oracle
% in Fig.~\ref{fig:oracle_combined}, reveal losses from both constraints.
% Table~\ref{tab:grouping-granularity} further shows the benefits of finer KV
% granularity and grouping more similar queries at matched realized densities.

Structural binding error arises from two constraints of block-sparse attention. \emph{KV binding} forces a block-sparse selector to retain irrelevant neighbors of important tokens, while \emph{query binding} forces queries with different attention patterns to share one selection (Fig.~\ref{fig:gap_sources}(a)). Moving from the block oracle to the token oracle with shared query blocks removes KV binding, and moving further to the per-query oracle removes query binding. The accuracy gains at each step (Fig.~\ref{fig:oracle_combined}) show that both constraints cause losses. Table~\ref{tab:grouping-granularity} further confirms this at matched realized densities: finer KV granularity and grouping more similar queries both improve fidelity.

\textbf{Token-level KV selection.}
To remove KV binding, MC-Sparse selects individual KV tokens for each query
group across block boundaries. The token level sparse attention kernel gathers the selected keys and
values online from their indices without repacking them into blocks.

\textbf{Tile-aligned query grouping.}
With head-specific keys and values in MHA, efficient attention tiles must
batch queries within each head. We therefore form groups
$\mathcal G_g$ of $C$ similar queries that share one KV selection, with $C$
matched to the kernel's query tile size. Minimizing within-group variance of
$\mathbf Q$ is equivalent to maximizing between-group variance.
Standard clustering methods such as $k$-means do not guarantee equal-size
groups, while enforcing hard balance constraints adds assignment overhead
and complicates GPU parallelization. We adopt a median-split version of
principal direction divisive partitioning
(PDDP)~\citep{boley98principal}: it recursively projects each group onto its
principal direction and splits the ordered queries at the median until every
group contains exactly $C$ queries, promoting within-group similarity while
maintaining tile alignment. Because a naive PDDP implementation is expensive
on GPUs, we
develop Fast PDDP as described in Section~\ref{sec:pipeline}.

% We also evaluated a key-covariance transformation before grouping, but it
% did not yield a significant improvement in end-to-end metrics. This negative
% result is better reported in the appendix or ablation section.

\subsection{Mitigating Selection Error}

% \textbf{Selection error.}
For a query group $\mathcal G_g$ and a budget of $K$ KV tokens, we define
\emph{oracle selection} as the set that maximizes the group's cumulative
dense-attention mass:
\begin{equation}
    \mathcal S_g^*
    = \underset{|\mathcal S|=K}{\arg\max}
    \sum_{i\in\mathcal G_g}\sum_{j\in\mathcal S} A_{ij}.
    \label{eq:oracle-selection}
\end{equation}
Here $\mathcal S$ is a set of KV indices. This objective is motivated by an
output-error bound proportional to discarded attention mass for bounded
value norms~\citep{shen2026ssasparsesparseattention}. Vanilla BSA instead scores pooled representations;
even pooling queries alone does not preserve average attention probabilities:
\begin{equation}
    \operatorname{Softmax}(\bar{\mathbf q}_g\mathbf K^\top/\sqrt d)_j
    \neq \frac{1}{|\mathcal G_g|}\sum_{i\in\mathcal G_g} A_{ij},
\end{equation}
where $\bar{\mathbf q}_g$ is the group mean. Pooled scoring can
miss KV tokens with high cumulative mass.
% SSA is the citation key supplied by the author; confirm the bibliography
% entry and its bound assumptions. Mass optimality is not output-error optimality.

% wrapfig is loaded in the main manuscript preamble.
\begin{wrapfigure}{R}{0.5\textwidth}
    \centering
    \includegraphics[width=\linewidth]{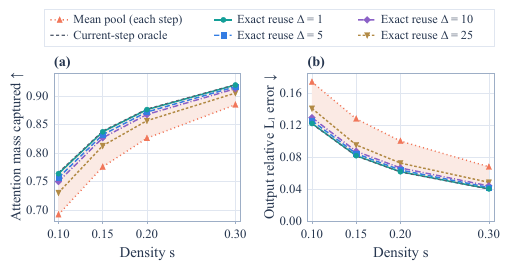}
    \caption{Temporal stability and accuracy of reused exact selection.}
    \label{fig:exact_selection_reuse}
\end{wrapfigure}
\textbf{Reusable exact KV Selection.}
Without a trained indexer, recomputing exact attention scores at every step is costly. Attention patterns remain stable across nearby denoising
steps~\citep{xia2025trainingfreeadaptivesparseattention}; exact selections from
earlier steps outperform mean-pooled selection recomputed at every step
(Fig.~\ref{fig:exact_selection_reuse}). This supports computing
Eq.~\ref{eq:oracle-selection} only at anchor steps and reusing the result.
At step $\tau$, we obtain $\mathcal S_{\tau,g}$ by selecting the top-$K$ tokens ranked by $\sum_{i\in\mathcal G_{\tau,g}} A_{\tau,ij}$. We cache the query groups
$\mathcal G_\tau$ and selected indices $\mathcal S_\tau$ for subsequent steps,
amortizing both grouping and selection. The current $\mathbf Q_t$,
$\mathbf K_t$ and $\mathbf V_t$ are recomputed at each step.

\subsection{Mitigating Discarded Tail Error}

\Needspace{20\baselineskip}
\begin{wrapfigure}{r}{0.48\textwidth}
    \vspace{-\intextsep}
    \centering
    \includegraphics[width=\linewidth]{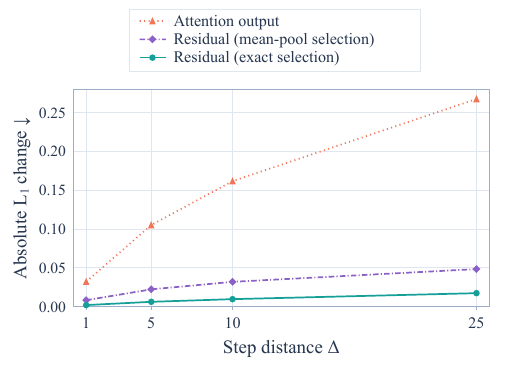}
    \caption{Absolute temporal change of attention outputs and dense--sparse
    residuals under token-level selection. Lower values indicate smaller
    absolute variation across $\Delta$ denoising steps.}
    \label{fig:residual_stability}
\end{wrapfigure}

% \textbf{Discarded tail error.}
Discarded tokens typically have small attention weights but still make nonzero
contributions (Fig.~\ref{fig:gap_sources}(c)), leaving an output gap to dense
attention even with per-query oracle selection.

\textbf{Temporal residual compensation.}
Block-sparse methods approximate discarded contributions by incorporating
block-mean representations into attention or correcting outputs with
block-level statistics~\citep{li2026pisapiecewisesparseattention,zhou2026svgearparameterfreelinearcompensation}.
Token-level selection cuts irregularly across these blocks, preventing direct
reuse of their summaries. We instead exploit the temporal dimension: the
dense--sparse residual varies less across denoising steps than the attention
output, with exact selection further reducing this variation compared with
mean-pooled selection (Fig.~\ref{fig:residual_stability}).
% Replace SVG-EAR with the manuscript's actual bibliography key.

We use a simple additive residual correction~\citep{yuan2024ditfastattnattentioncompressiondiffusion}.
At anchor step $\tau$, we record
\begin{equation}
\mathbf{R}_{\tau}
=
\mathbf{O}^{\mathrm{full}}_{\tau}
-
\mathbf{O}^{\mathrm{sparse}}_{\tau},
\end{equation}
where $\mathbf{O}^{\mathrm{sparse}}_{\tau}$ uses the KV selection. At a
subsequent reuse step $t$, MC-Sparse computes
\begin{equation}
\widehat{\mathbf{O}}_t
=
\operatorname{SparseAttn}
\left(
\mathbf{Q}_t,\mathbf{K}_t,\mathbf{V}_t;
\mathcal{G}_{\tau},\mathcal{S}_{\tau}
\right)
+
\mathbf{R}_{\tau}.
\end{equation}
The residual follows the same refresh schedule as the cached groups and KV
indices. Because the anchor step already computes dense attention for exact
selection, forming the residual adds one sparse-attention evaluation and an
elementwise subtraction at each anchor.

% Our residual compensation yields larger gains on vanilla BSA than on PISA
% (Table~\ref{tab:compensation-across-methods}). Both our compensation and
% PISA's native block-statistics correction serve a similar role in
% compensating for discarded tail contributions. This shared role suggests
% that PISA's correction becomes less critical when our compensation is applied.
\subsection{Pipeline and Efficient Implementation}
\label{sec:pipeline}

\begin{figure}[!t]
    \centering
    \captionsetup{skip=3pt}
    \includegraphics[width=\linewidth]{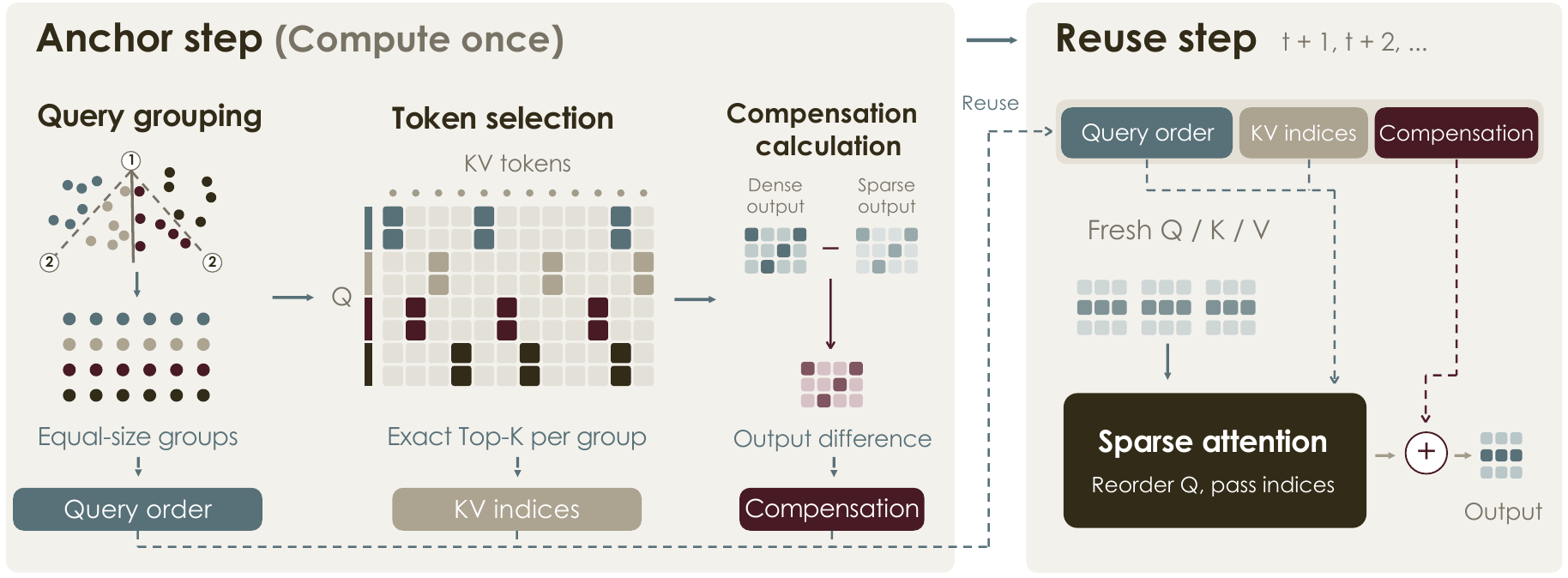}
    \caption{Pipeline overview of MC-Sparse.}
    \label{fig:pipeline}
\end{figure}

\textbf{Overall pipeline.}
MC-Sparse alternates between anchor and reuse steps. At an anchor step, which
is the last step of a dense warm-up or refresh segment, we run dense attention,
construct query groups, perform exact KV selection, and cache the query groups,
selected indices, and output residual. At each subsequent reuse step, the
current $\mathbf{Q}$, $\mathbf{K}$, and $\mathbf{V}$ are recomputed, while
attention uses the cached metadata and residual. Figure~\ref{fig:pipeline}
illustrates the pipeline; Algorithm~\ref{alg:mcsparse} gives the corresponding
pseudocode.

% Algorithm block included from the main Method section.
\begin{algorithm}[!htbp]
\caption{MC-Sparse}
\label{alg:mcsparse}
\small
\begin{algorithmic}[1]
\REQUIRE Diffusion steps $T$, reuse interval $L$, KV budget $K$
\ENSURE Attention outputs $\{\widehat{\mathbf{O}}_t\}_{t=1}^{T}$
\FOR{$t=1,\ldots,T$}
    \STATE Compute $\mathbf{Q}_t$, $\mathbf{K}_t$, and $\mathbf{V}_t$
    \IF{$\textsc{IsAnchor}(t,L)$}
        \STATE $(\mathbf{O}^{\mathrm{full}}_t,\mathbf{l}_t)
        \leftarrow \textsc{FlashAttn}
        (\mathbf{Q}_t,\mathbf{K}_t,\mathbf{V}_t)$
        \STATE $\mathcal{G}_t\leftarrow\textsc{GroupQueries}(\mathbf{Q}_t)$
        \STATE $\mathcal{S}_t\leftarrow\textsc{ExactKVSelection}
        (\mathbf{Q}_t,\mathbf{K}_t,\mathbf{l}_t,\mathcal{G}_t,K)$
        \STATE $\mathbf{O}^{\mathrm{sparse}}_t\leftarrow\textsc{SparseAttn}
        (\mathbf{Q}_t,\mathbf{K}_t,\mathbf{V}_t;
        \mathcal{G}_t,\mathcal{S}_t)$
        \STATE $\mathbf{R}_t\leftarrow
        \mathbf{O}^{\mathrm{full}}_t-\mathbf{O}^{\mathrm{sparse}}_t$
        \STATE $\mathcal{C}\leftarrow
        (\mathcal{G}_t,\mathcal{S}_t,\mathbf{R}_t)$
        \STATE $\widehat{\mathbf{O}}_t\leftarrow
        \mathbf{O}^{\mathrm{full}}_t$
    \ELSE
        \STATE $(\mathcal{G},\mathcal{S},\mathbf{R})\leftarrow\mathcal{C}$
        \STATE $\mathbf{O}^{\mathrm{sparse}}_t\leftarrow\textsc{SparseAttn}
        (\mathbf{Q}_t,\mathbf{K}_t,\mathbf{V}_t;\mathcal{G},\mathcal{S})$
        \STATE $\widehat{\mathbf{O}}_t\leftarrow
        \mathbf{O}^{\mathrm{sparse}}_t+\mathbf{R}$
    \ENDIF
\ENDFOR
\end{algorithmic}
\end{algorithm}

\textbf{Fast PDDP.}
We obtain principal directions using batched power iteration, processing all nodes at the same tree depth across attention heads in parallel. To reduce peak memory usage and improve speed, we propagate only query-index arrays between tree levels. Table \ref{tab:grouping-cost} reports its single-call and amortized grouping costs.

\textbf{Two-pass exact selection.}
FlashAttention does not expose attention logits, so we implement a two-pass
kernel for exact KV selection. Details are provided in Appendix~\ref{app:implementation_details}.

\textbf{Token-sparse attention kernel.}
We implement the token-sparse attention kernel in CuTeDSL, with performance
comparisons in Table~\ref{tab:attn-kernel-speed} and design details in
Appendix~\ref{app:implementation_details}.

\textbf{Cache offloading and prefetching.}
When GPU memory is insufficient, we offload inactive cache entries to CPU
memory and prefetch them ahead of use, overlapping transfers with computation to hide latency. Table~\ref{tab:offload} demonstrates the effectiveness of prefetching.
\FloatBarrier
% Required packages: booktabs, graphicx, wrapfig, natbib, xcolor (with table).
% Standalone expanded main-experiment section. Include the separate ablation
% and system-efficiency sections after this material.
\section{Experiments}
\definecolor{lightrowblue}{HTML}{EAF2F8}
\definecolor{graydivider}{HTML}{F2F4F4}

\subsection{Experimental Setup}
\textbf{Tasks and baselines.}
We evaluate video generation with Minimax-H3-Base~\citep{minimax2026h3},
HunyuanVideo-13B~\citep{kong2025hunyuanvideosystematicframeworklarge}, and
Wan2.1-14B-T2V/I2V~\citep{wan2025wanopenadvancedlargescale}, and image-to-geometry
generation with HY3D-Internal. Following
SVG-EAR~\citep{zhou2026svgearparameterfreelinearcompensation}, we use Penguin
Benchmark prompts for text-to-video and VBench image--prompt pairs for
image-to-video, and a detail-oriented image test set for 3D generation.
We compare with FA3-based dense attention~\citep{shah2024flashattention3fastaccurateattention},
SVG2, SVG-EAR, Sol-Attn~\citep{li2026solattnacceleratingvideogeneration},
PISA~\citep{li2026pisapiecewisesparseattention}, and
SpargeAttn~\citep{zhang2025spargeattentionaccuratetrainingfreesparse},
as applicable to each model.
% TODO: Add the existing SVG2 bibliography key.

\textbf{Metrics.}
PSNR, SSIM, and LPIPS~\citep{zhang2018unreasonableeffectivenessdeepfeatures}
measure video fidelity to dense-attention outputs, while VBench~\citep{huang2023vbenchcomprehensivebenchmarksuite}
measures image quality (ImgQual) and background consistency (BgCons).
For 3D assets, we report geometric fidelity using Chamfer distance (CD),
volumetric IoU, and F1, and input-image consistency using
Uni3D-I~\citep{zhou2023uni3dexploringunified3d} and
ULIP3D-I~\citep{xue2023uliplearningunifiedrepresentation}.
Efficiency is measured by attention density, the fraction of retained
query--key interactions, and DiT denoising speedup over dense attention,
including warm-up but excluding condition encoding and VAE decoding.
Detailed settings are provided in Appendix~\ref{app:experimental_details}.
\subsection{Main Results}
\begin{figure}[!htbp]
    \centering
    \includegraphics[width=\linewidth]{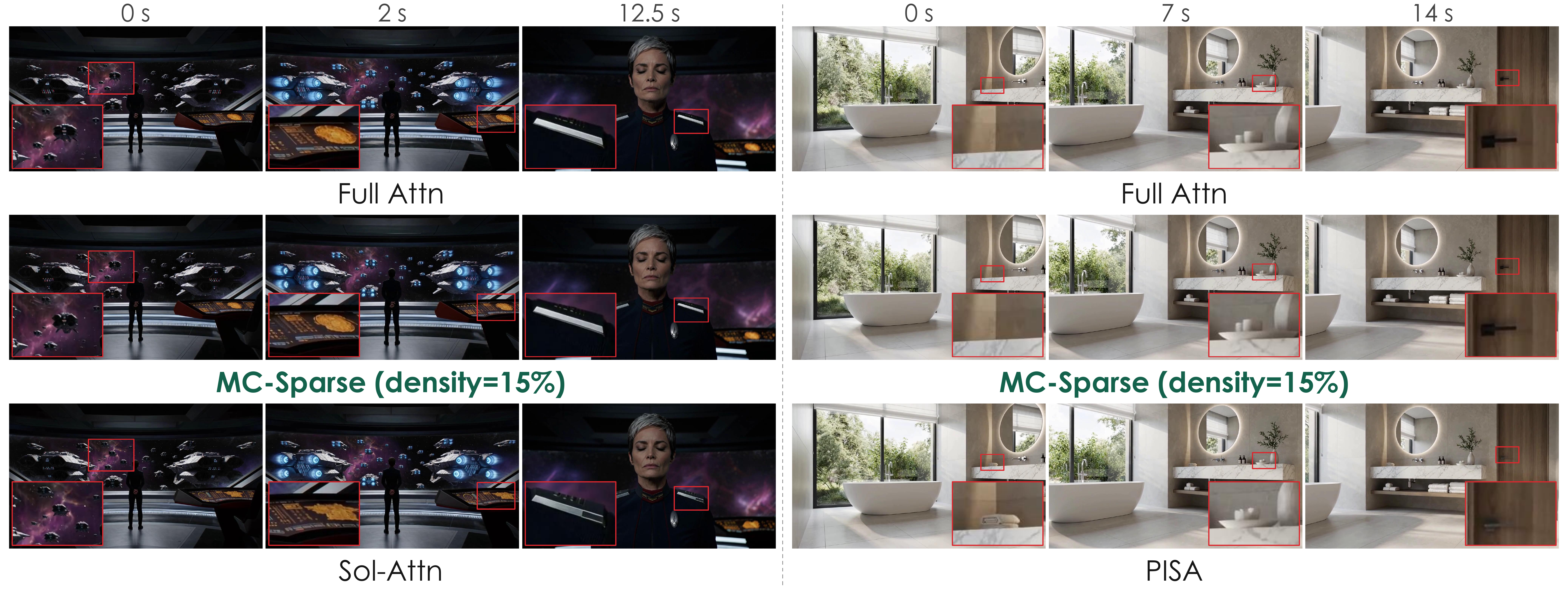}
    \caption{Qualitative comparison of video generation using Minimax-H3-Base.}
    \label{fig:video_qualitative}
\end{figure}

\textbf{Video generation.}
Table~\ref{tab:performance_comparison} shows improved fidelity and denoising
speedup across the evaluated video models. On Minimax-H3-Base, MC-Sparse
achieves 28.44 dB PSNR at 25\% density and $1.61\times$ speedup, compared with
23.66 dB and $1.59\times$ for Sol-Attn at 32.4\% density.
With the same anchor schedule, MC-Sparse-Flash reaches $1.80\times$ speedup
at 15\% density with 27.30 dB
PSNR, retaining higher fidelity than both Sol-Attn and PISA.
On Hunyuan and Wan, MC-Sparse achieves the highest PSNR and SSIM and the
lowest LPIPS among the evaluated sparse methods, with
$1.52\times$--$1.82\times$ speedup.
The qualitative comparisons in Fig.~\ref{fig:video_qualitative} show that
MC-Sparse produces videos visually closer to the dense-attention outputs
than the compared sparse methods, consistent with the fidelity metrics.
Qualitative results on other video generation models are provided in Fig.~\ref{fig:supp_qualitative_comparison}.

\begin{table}[!htbp]
\centering
\caption{Video generation results. PSNR, SSIM, and LPIPS compare with
dense-attention outputs; ImgQual and BgCons are VBench scores.
Speedup measures DiT denoising only.
Bold and underlined values indicate the best and second-best results,
respectively, excluding Full Attn. }
\label{tab:performance_comparison}
\resizebox{\textwidth}{!}{
\begin{tabular}{l|ccccc|cc}
\toprule
\textbf{Method}
& \textbf{PSNR}$\uparrow$
& \textbf{SSIM}$\uparrow$
& \textbf{LPIPS}$\downarrow$
& \textbf{ImgQual}$\uparrow$
& \textbf{BgCons}$\uparrow$
& \textbf{Density}$\downarrow$
& \textbf{Speedup}$\uparrow$ \\
\midrule

\multicolumn{8}{l}{\textbf{Minimax-H3-Base, 768p, T2AV (video branch)}} \\
\midrule
Full Attn (FA3)
& - & - & -
& 69.20
& 91.72
& 100\% & $1.00\times$ \\

Sol-Attn
& 23.66 & 0.816 & 0.234
& 68.83
& 91.73
& 32.4\% & $1.59\times$ \\

PISA
& 23.45 & 0.811 & 0.243
& 68.54
& 91.64
& 30.0\% & $1.52\times$ \\

\rowcolor{lightrowblue}
\textbf{MC-Sparse (density=25\%)}
& \textbf{28.44} & \textbf{0.899} & \textbf{0.161}
& \textbf{69.00}
& \underline{91.81}
& 25.0\% & $\underline{1.61\times}$ \\

\rowcolor{lightrowblue}
\textbf{MC-Sparse (density=15\%)}
& \underline{27.30} & \underline{0.882} & \underline{0.176}
& \underline{68.94}
& \textbf{91.85}
& 15.0\% & $\mathbf{1.80\times}$ \\
\midrule

\multicolumn{8}{l}{\textbf{HunyuanVideo-13B, 720p, T2V}} \\
\midrule
Full Attn (FA3)
& - & - & -
& 68.52
& 96.91
& 100\% & $1.00\times$ \\

Sol-Attn
& 28.06 & 0.898 & 0.159
& 68.27
& \textbf{97.03}
& 32.3\% & $1.79\times$ \\

PISA
& 27.65 & 0.890 & 0.174
& \textbf{68.74}
& \underline{96.96}
& 30.0\% & $1.70\times$ \\

\rowcolor{lightrowblue}
\textbf{MC-Sparse (density=25\%)}
& \textbf{32.89} & \textbf{0.940} & \textbf{0.114}
& 68.66
& 96.89
& 25.0\% & $\underline{1.82\times}$ \\

\rowcolor{lightrowblue}
\textbf{MC-Sparse (density=15\%)}
& \underline{30.78} & \underline{0.919} & \underline{0.135}
& \underline{68.67}
& \underline{96.96}
& 15.0\% & $\mathbf{2.00\times}$ \\
\midrule

\multicolumn{8}{l}{\textbf{Wan2.1-14B-T2V, 720p}} \\
\midrule
Full Attn (FA3)
& - & - & -
& 67.11
& 96.60
& 100\% & $1.00\times$ \\

SpargeAttn
& 23.56 & 0.828 & 0.212
& 66.84
& \textbf{96.81}
& 30.0\% & - \\

Sol-Attn
& 24.62 & 0.852 & 0.157
& 67.01
& 96.72
& 31.6\% & $\underline{1.52\times}$ \\

PISA
& 24.03 & 0.838 & 0.202
& 67.09
& \underline{96.76}
& 30.0\% & $\underline{1.52\times}$ \\

SVG2
& 26.02 & 0.875 & 0.166
& \underline{67.11}
& 96.52
& 31.5\% & $1.34\times$ \\

SVG-EAR
& \underline{27.61} & \underline{0.897} & \underline{0.142}
& 66.97
& 96.41
& 25.3\% & $1.38\times$ \\

\rowcolor{lightrowblue}
\textbf{MC-Sparse (Ours)}
& \textbf{28.81} & \textbf{0.912} & \textbf{0.128}
& \textbf{67.12}
& 96.70
& 25.0\% & $\mathbf{1.53\times}$ \\
\midrule

\multicolumn{8}{l}{\textbf{Wan2.1-14B-I2V, 720p}} \\
\midrule
Full Attn (FA3)
& - & - & -
& 70.37
& 96.51
& 100\% & $1.00\times$ \\

SpargeAttn
& 26.66 & 0.859 & 0.172
& \underline{70.40}
& 96.30
& 30.0\% & - \\

Sol-Attn
& 27.73 & 0.876 & 0.157
& 70.34
& \textbf{96.46}
& 31.9\% & $\underline{1.51\times}$ \\

PISA
& 27.10 & 0.865 & 0.163
& 70.38
& 96.40
& 30.0\% & $\underline{1.51\times}$ \\

SVG2
& 27.15 & 0.861 & 0.168
& 70.32
& \underline{96.42}
& 30.4\% & $1.35\times$ \\

SVG-EAR
& \underline{30.71} & \underline{0.916} & \underline{0.127}
& 70.34
& 96.38
& 24.6\% & $1.39\times$ \\

\rowcolor{lightrowblue}
\textbf{MC-Sparse (Ours)}
& \textbf{32.11} & \textbf{0.929} & \textbf{0.117}
& \textbf{70.41}
& \textbf{96.46}
& 25.0\% & $\mathbf{1.52\times}$ \\

\bottomrule
\end{tabular}
}
\end{table}

\begin{figure}[!htbp]
    \centering
    \includegraphics[width=\linewidth]{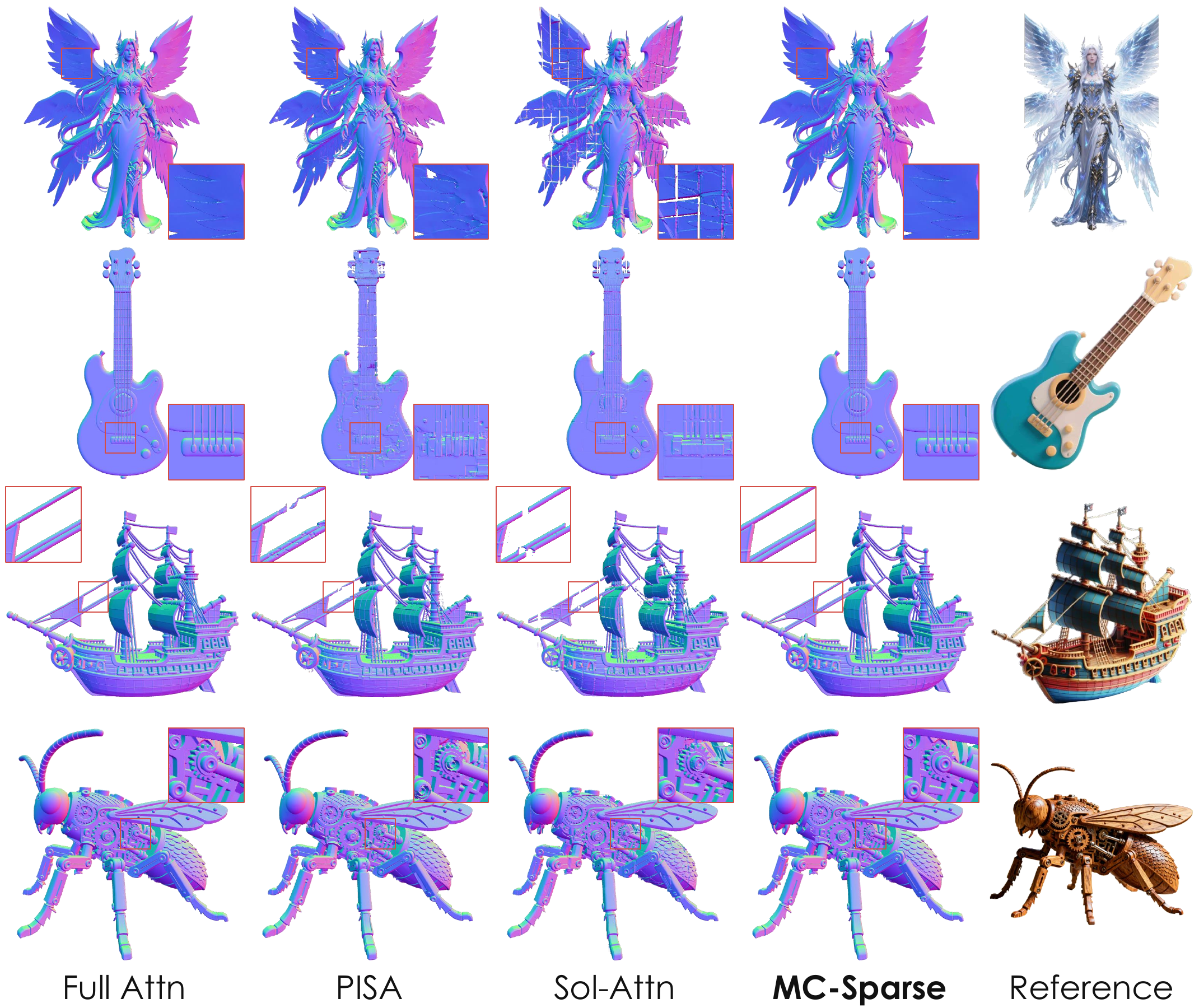}
    \caption{Qualitative comparison on 3D asset generation.}
    \label{fig:3d_qualitative}
\end{figure}

\textbf{3D asset generation.}
At 15\% attention density, MC-Sparse achieves $2.32\times$ denoising speedup
and the best geometric fidelity among the evaluated sparse methods
(Table~\ref{tab:hy3d_geometry_comparison}), reducing CD from PISA's 0.976
to 0.177. Its input-image consistency remains comparable to dense attention.
In Fig.~\ref{fig:3d_qualitative}, MC-Sparse better preserves fine details and
surface continuity, while the compared methods exhibit broken surfaces and
missing details despite using higher attention densities.

\begin{table}[!htbp]
\centering
\caption{Image-to-geometry results on HY3D-Internal.
Geometric fidelity is measured against dense-attention outputs; Uni3D-I and
ULIP3D-I measure input-image consistency. Speedup measures DiT denoising only.}
\label{tab:hy3d_geometry_comparison}
\resizebox{\textwidth}{!}{
\begin{tabular}{l|ccccc|cc}
\toprule
\textbf{Config} & \textbf{CD}$\downarrow$ & \textbf{Vol-IoU-1536}$\uparrow$ & \textbf{F1@0.001}$\uparrow$ & \textbf{Uni3D-I}$\uparrow$ & \textbf{ULIP3D-I}$\uparrow$ & \textbf{Density}$\downarrow$ & \textbf{Speedup}$\uparrow$ \\ 
\midrule
\multicolumn{8}{l}{\textbf{HY3D-Internal}} \\
\midrule
Full Attn (FA3) & - & - & - &  0.3309 & 0.1204  & 100\% & $1.00\times$ \\
Sol-Attn       & 1.901 & 50.55 & 70.36 & 0.3301 & 0.1203 & 36.1\% & $1.57\times$ \\
PISA       & 0.976  & 63.22  & 83.29 & 0.3304  & 0.1204 & 25.0\% &  $<1.87\times$\\
\rowcolor{lightrowblue} \textbf{MC-Sparse (Ours)} & \textbf{0.177}  & \textbf{82.91} &  \textbf{96.33} & \textbf{0.3311} & \textbf{0.1206} & 15.0\% & $\mathbf{2.32\times}$  \\
\bottomrule
\end{tabular}
}
\end{table}

\subsection{Ablation Study}
\label{sec:ablation}

\textbf{Component contributions.}
Starting from vanilla BSA, we successively add exact KV selection with
cross-step reuse, token-level KV granularity, query grouping, and residual compensation (Table~\ref{tab:mcsparse-ablation}). Each addition improves PSNR, SSIM, and LPIPS across all evaluated densities, supporting the contribution of each component to the full framework.

\begin{table}[!htbp]
    \centering
    \caption{Cumulative ablation on Wan2.1-1.3B-T2V at 480p. Each row adds
    one component to the preceding configuration. Settings are provided in
    Appendix~\ref{app:experimental_details}.}
    \label{tab:mcsparse-ablation}
    \small
    \setlength{\tabcolsep}{4pt}
    \renewcommand{\arraystretch}{1.12}
    \resizebox{\textwidth}{!}{%
    \begin{tabular}{@{}l ccc ccc ccc@{}}
        \toprule
        \multirow{2}{*}{Configuration}
        & \multicolumn{3}{c}{Density $s=0.2$}
        & \multicolumn{3}{c}{Density $s=0.25$}
        & \multicolumn{3}{c}{Density $s=0.3$} \\
        \cmidrule(lr){2-4}\cmidrule(lr){5-7}\cmidrule(lr){8-10}
        & PSNR $\uparrow$ & SSIM $\uparrow$ & LPIPS $\downarrow$
        & PSNR $\uparrow$ & SSIM $\uparrow$ & LPIPS $\downarrow$
        & PSNR $\uparrow$ & SSIM $\uparrow$ & LPIPS $\downarrow$ \\
        \midrule
        Vanilla BSA
        & 20.96 & 0.7535 & 0.2714
        & 21.67 & 0.7750 & 0.2469
        & 22.26 & 0.7923 & 0.2278 \\
        $+$ exact KV selection (w. reuse)
        & 21.60 & 0.7731 & 0.2482
        & 22.63 & 0.8004 & 0.2176
        & 23.47 & 0.8205 & 0.1956 \\
        $+$ token granularity
        & 22.57 & 0.7954 & 0.2254
        & 23.54 & 0.8188 & 0.1995
        & 24.38 & 0.8380 & 0.1789 \\
        $+$ query grouping
        & 23.51 & 0.8175 & 0.2017
        & 24.47 & 0.8386 & 0.1784
        & 25.24 & 0.8539 & 0.1624 \\
        $+$ compensation (full model)
        & \textbf{27.05} & \textbf{0.8809} & \textbf{0.1326}
        & \textbf{27.35} & \textbf{0.8843} & \textbf{0.1294}
        & \textbf{27.54} & \textbf{0.8862} & \textbf{0.1273} \\
        \bottomrule
    \end{tabular}
    }
\end{table}

\textbf{KV granularity and query grouping.}
Table~\ref{tab:grouping-granularity} shows that token-level KV selection
improves attention recall and output accuracy under every grouping strategy.
We match realized density after tile alignment, since padding variable-size
$k$-means groups increases executed interactions beyond the nominal budget.
Under this comparison, Fast PDDP avoids alignment inflation and consistently
outperforms $k$-means, supporting both token-level selection and tile-aligned
query grouping.

\begin{table}[!htbp]
\centering\small
\setlength{\tabcolsep}{4pt}
\renewcommand{\arraystretch}{1.08}
\caption{KV-selection granularity and query grouping at matched realized attention density $s$. \emph{Align.} is the ratio of executed to nominal sparse interactions after tile alignment. Black rows include this inflation in the density budget, whereas gray rows ignore it$^{\dagger}$.}
\label{tab:grouping-granularity}
\label{tab:grouping-granularity-full}
\begin{tabular}{llccccccc}
\toprule
\multirow{2}{*}{Query grouping} & \multirow{2}{*}{KV granularity} & \multirow{2}{*}{Align.} & \multicolumn{3}{c}{Attention recall $\uparrow$} & \multicolumn{3}{c}{Output rel. $L_1$ $\downarrow$} \\
\cmidrule(lr){4-6}\cmidrule(lr){7-9}
& & & $s{=}0.1$ & $s{=}0.2$ & $s{=}0.3$ & $s{=}0.1$ & $s{=}0.2$ & $s{=}0.3$ \\
\midrule
\multirow{2}{*}{No reorder} & block & 1.00 & 0.637 & 0.778 & 0.847 & 0.176 & 0.101 & 0.069 \\
 & token & 1.00 & 0.745 & 0.857 & 0.910 & 0.130 & 0.070 & 0.044 \\
\addlinespace[2pt]
\multirow{4}{*}{k-means} & block & 2.08 & 0.743 & 0.830 & 0.874 & 0.160 & 0.098 & 0.071 \\
 & token & 1.45 & 0.840 & 0.904 & 0.935 & 0.093 & 0.054 & 0.036 \\
& \textcolor{gray}{\itshape block$^{\dagger}$} & \textcolor{gray}{2.08} & \textcolor{gray}{0.834} & \textcolor{gray}{0.906} & \textcolor{gray}{0.940} & \textcolor{gray}{0.096} & \textcolor{gray}{0.052} & \textcolor{gray}{0.033} \\
& \textcolor{gray}{\itshape token$^{\dagger}$} & \textcolor{gray}{1.45} & \textcolor{gray}{0.876} & \textcolor{gray}{0.933} & \textcolor{gray}{0.959} & \textcolor{gray}{0.071} & \textcolor{gray}{0.038} & \textcolor{gray}{0.023} \\
\addlinespace[2pt]
\multirow{2}{*}{\textbf{Fast PDDP (ours)}} & block & 1.00 & 0.791 & 0.887 & 0.930 & 0.118 & 0.062 & 0.038 \\
 & \textbf{token} & 1.00 & \textbf{0.856} & \textbf{0.923} & \textbf{0.953} & \textbf{0.080} & \textbf{0.042} & \textbf{0.026} \\
\midrule
\textit{Oracle (per query)} & token & -- & 0.908 & 0.951 & 0.971 & 0.048 & 0.025 & 0.015 \\
\bottomrule
\end{tabular}
\end{table}

\textbf{Role of residual compensation.}
Residual compensation yields the largest gain in the cumulative ablation,
but its benefit is substantially greater with MC-Sparse's selection and
grouping designs than with vanilla BSA
(Table~\ref{tab:compensation-across-methods}), indicating that its effectiveness
depends on their combination. PISA receives smaller additional gains than
vanilla BSA with its native block-statistics compensation retained, consistent
with both mechanisms compensating for discarded tail contributions.

% Required package: booktabs.

\begin{table}[!htbp]
    \centering
    \caption{PSNR (dB) before and after adding our residual compensation at
    target attention density $s$. $\Delta$ is the PSNR gain from adding the
    residual. PISA retains its native block-statistics compensation in both
    configurations.}
    \label{tab:compensation-across-methods}
    \small
    \setlength{\tabcolsep}{4pt}
    \renewcommand{\arraystretch}{1.12}
    \begin{tabular}{@{}l cc cc cc@{}}
        \toprule
        & \multicolumn{2}{c}{$s=0.20$}
        & \multicolumn{2}{c}{$s=0.25$}
        & \multicolumn{2}{c}{$s=0.30$} \\
        \cmidrule(lr){2-3}\cmidrule(lr){4-5}\cmidrule(lr){6-7}
        Configuration & PSNR $\uparrow$ & $\Delta$
        & PSNR $\uparrow$ & $\Delta$ & PSNR $\uparrow$ & $\Delta$ \\
        \midrule
        MC-Sparse w/o residual compensation
        & 23.51 & -- & 24.47 & -- & 25.24 & -- \\
        MC-Sparse
        & \textbf{27.05} & \textbf{+3.54}
        & \textbf{27.35} & \textbf{+2.88}
        & \textbf{27.54} & \textbf{+2.30} \\
        \midrule
        Vanilla BSA & 20.96 & -- & 21.67 & -- & 22.26 & -- \\
        \quad $+$ our residual compensation
        & 22.21 & +1.25 & 23.12 & +1.45 & 23.90 & +1.64 \\
        \midrule
        PISA & 21.66 & -- & 22.14 & -- & 22.52 & -- \\
        \quad $+$ our residual compensation
        & 22.15 & +0.49 & 22.92 & +0.78 & 23.42 & +0.90 \\
        \bottomrule
    \end{tabular}
\end{table}

\Needspace{8\baselineskip}
\subsection{System Efficiency}
\label{sec:system-efficiency}

\textbf{Token-sparse attention kernel.}
Table~\ref{tab:attn-kernel-speed} compares attention kernels at matched
KV-token densities on Wan2.1-14B-T2V at 720p, excluding selection and grouping.
SVG2 uses $Q_c=300$ and $K_c=1000$. MC-Sparse achieves 96\%--99\% of the ideal
$1/s$ speedup, versus 80\%--81\% for SVG2 and 82\%--87\% for PISA, showing
that token-level KV gathering preserves kernel efficiency.

\begin{table}[H]
% \vspace{-3pt}
\centering
\small
\setlength{\tabcolsep}{4pt}
\renewcommand{\arraystretch}{1.08}
\caption{Attention-kernel speedup on Wan2.1-14B-T2V (720p), using a Hopper GPU in BF16 ($H=40$, $D=128$, $S=75600$).
Entries show speedup $\rho=t_{\mathrm{FA3}}/t$ (efficiency $\eta=\rho s$ relative to the ideal $1/s$ speedup).
Selection and grouping are excluded.}
\label{tab:attn-kernel-speed}
\begin{tabular}{@{}llccc@{}}
\toprule
Kernel & Sparse layout & $s=0.2$ & $s=0.3$ & $s=0.5$ \\
\midrule
\textcolor{gray}{BSA} & \textcolor{gray}{regular blocks} & \textcolor{gray}{4.96$\times$ (0.99)} & \textcolor{gray}{3.30$\times$ (0.99)} & \textcolor{gray}{1.97$\times$ (0.99)} \\
PISA kernel & regular blocks & 4.11$\times$ (0.82) & 2.84$\times$ (0.85) & 1.73$\times$ (0.87) \\
SVG2 kernel & variable-length blocks & 4.02$\times$ (0.80) & 2.70$\times$ (0.81) & 1.62$\times$ (0.81) \\
MC-Sparse & tile-aligned Q, gathered KV & 4.79$\times$ (\textbf{0.96}) & 3.25$\times$ (\textbf{0.98}) & 1.97$\times$ (\textbf{0.99}) \\
\bottomrule
\end{tabular}
\end{table}

\Needspace{16\baselineskip}
\begin{wraptable}{r}{0.50\textwidth}

        \vspace{-\intextsep}
        \centering
        \caption{Query-grouping latency on Wan2.1-T2V (ms per layer).
        Amortized over 40 steps: 2 Fast PDDP calls or Flash-KMeans with a 50-iteration initialization and 39 two-iteration updates.}
        \label{tab:grouping-cost}
        \small
        \setlength{\tabcolsep}{2pt}
        \renewcommand{\arraystretch}{1.08}
        \begin{tabular}{@{}lrrrr@{}}
            \toprule
            & \multicolumn{2}{c}{1.3B, 480p} & \multicolumn{2}{c}{14B, 720p} \\
            \cmidrule(lr){2-3}\cmidrule(lr){4-5}
            Grouping & Single & Amort. & Single & Amort. \\
            \midrule
            Flash-KMeans init.
            & 32.0 & \multirow{2}{*}{2.26} & 295.1 & \multirow{2}{*}{20.43} \\
            \quad $+$ step update & 1.50 & & 13.38 & \\
            \addlinespace[2pt]
            \textbf{Fast PDDP}
            & 11.3 & \textbf{0.57} & 66.6 & \textbf{3.33} \\
            \bottomrule
        \end{tabular}

\end{wraptable}

{\raggedright
\textbf{Tile-aligned query grouping.}
Table~\ref{tab:grouping-cost} compares Fast PDDP with Flash-KMeans.
Under the evaluated schedules, Fast PDDP reduces amortized grouping cost by
$3.96\times$ at 480p and $6.14\times$ at 720p, keeping tile-aligned query
grouping inexpensive.

\textbf{Anchor-step overhead.}
The two-pass selector adds a QK-only pass costing approximately half a dense
attention evaluation in GEMM operations. With the first anchor overlapping
the final warm-up step (Table~\ref{tab:appendix-warmup}), the additional dense
evaluation and selection passes yield an amortized overhead equivalent to an
absolute density increase of $\Delta s\approx5\%$ over all denoising steps.
Including query grouping and
other auxiliary operations, the measured overhead is equivalent to
$\Delta s\approx5\%$--$6\%$, below SVG2's combined clustering and
variable-length kernel overhead.\par}

\textbf{Cache offloading and prefetching.}
Table~\ref{tab:offload} reports per-layer transfer and sparse-attention costs.
For both evaluated models, transferring the cached residuals and selected KV
indices takes less time than sparse-attention computation. Prefetching one
layer ahead hides the transfer latency in these measured settings.

\begin{table}[!htb]
    \centering
    \small
    \setlength{\tabcolsep}{4pt}
    \renewcommand{\arraystretch}{1.08}
    \caption{Per-layer cost of offloading cached output residuals and selected KV indices to CPU memory.
    $V$ is the transfer volume and $C$ the sparse-attention FLOPs; $C/V$ gives the break-even compute-to-PCIe-bandwidth ratio for overlap.
    Check marks indicate transfers hidden by prefetching one layer ahead in the measured setting.}
    \label{tab:offload}
    \begin{tabular}{@{}lrrrrrc@{}}
        \toprule
        Model & \shortstack{$V$\\(MB)} & \shortstack{$C$\\(TFLOP)}
        & \shortstack{$C/V$\\(FLOP/B)} & \shortstack{Transfer\\(ms)}
        & \shortstack{Compute\\(ms)} & \shortstack{Transfer\\hidden} \\
        \midrule
        Wan2.1-14B-T2V & 2435 & 29.1 & 11,409 & 47.7 & 208.8 & $\checkmark$ \\
        HunyuanVideo-13B & 2710 & 34.6 & 12,164 & 52.9 & 244.5 & $\checkmark$ \\
        \bottomrule
    \end{tabular}
\end{table}
\FloatBarrier
\section{Conclusion}

We presented MC-Sparse, a training-free framework for accelerating diffusion
transformers with token-level sparse attention. Our oracle comparisons show
that the dense-to-sparse gap extends beyond selection accuracy to include
structural binding and discarded tail contributions. Guided by this analysis,
MC-Sparse combines token-level KV selection and tile-aligned query grouping
with temporal reuse and residual compensation, balancing approximation
quality with execution efficiency. Experiments on video and 3D asset
generation demonstrate improved fidelity to dense-attention outputs and
reduced denoising latency compared with the evaluated sparse-attention
baselines.

\clearpage
\bibliography{iclr2027_conference}
\bibliographystyle{colm2024_conference}

\clearpage
\appendix
% AI Use Disclosure disabled at the author's request; source retained.
% \section{AI Use Disclosure}
% We used generative AI tools to assist with drafting and editing the manuscript,
% including feedback on the presentation of the methodology, experiments, and
% results, and to generate or revise LaTeX, plotting code, figures, and tables.
% The authors reviewed and verified all AI-assisted content. Generative AI was
% not used to generate experimental data or model outputs. The authors take full
% responsibility for the final content of this work.
%

% Required packages: booktabs, graphicx.
% TODO: Add existing bibliography keys for SVG2, SVG-EAR, VBench, and Penguin.

\section{Experimental Details}
\label{app:experimental_details}

\subsection{Evaluation Protocol}
\label{app:evaluation_protocol}

\paragraph{Video generation evaluation.}
Following SVG-EAR, we use the Penguin Benchmark text-to-video prompts refined
by the VBench team and the VBench image--prompt pairs with enhanced prompts for
image-to-video generation. We randomly select 50 samples from each task's
evaluation set. For Minimax-H3-Base, we further rewrite the test prompts with the official prompt-enhancement API and generate 345-frame videos, approximately 14.4
seconds long.

\paragraph{Analysis and ablation.}
We use Wan2.1-1.3B-T2V at 480p only for the analysis and ablation experiments.
When computing the analysis metrics, we uniformly sample diffusion steps and
transformer layers outside the dense warm-up regions.
Analysis experiments use the same evaluation set as the corresponding
generation task. We compute the reported metrics by uniformly sampling
diffusion steps and transformer layers outside the dense warm-up regions.

\paragraph{Warm-up configuration.}
Table~\ref{tab:appendix-warmup} lists the dense warm-up steps and layers for
the video model configurations and the 3D generation experiment. Each
entry gives the number of warm-up steps or layers divided by the corresponding
total. For Hunyuan, $1/20$ and $1/40$ refer to the dual-stream and single-stream
layers, respectively. MC-Sparse uses the same dense warm-up configuration as
the corresponding baseline; its anchor steps are also listed in the table.

\begin{table}[!htbp]
    \centering
    \caption{Dense warm-up and MC-Sparse anchor configurations. Warm-up entries
    denote count / total count. Hunyuan lists dual-stream and single-stream
    layers, respectively.}
    \label{tab:appendix-warmup}
    \small
    \setlength{\tabcolsep}{5pt}
    \renewcommand{\arraystretch}{1.12}
    \resizebox{\linewidth}{!}{%
    \begin{tabular}{@{}llccc@{}}
        \toprule
        Model & Setting & Warm-up layers & Warm-up steps & Anchor steps \\
        \midrule
        Wan2.1-1.3B-T2V & 480p, 5s, text-to-video & 1/30 & 10/50 & $\{10,26\}$ \\
        Wan2.1-14B-I2V & 720p, 5s, image-to-video & 1/40 & 10/50 & $\{10,26\}$ \\
        Wan2.1-14B-T2V & 720p, 5s, text-to-video & 1/40 & 10/50 & $\{10,26\}$ \\
        Hunyuan-13B & 720p, 5.25s, text-to-video & 1/20, 1/40 & 10/50 & $\{10,26\}$ \\
        Minimax-H3-Base & 768p, 14.4s, text-to-audio-video & 1/52 & 10/49 & $\{10,26\}$ \\
        \midrule
        HY3D-Internal & 1536 resolution, image-to-geometry & 1/40 & 1/12 & $\{1\}$ \\
        \bottomrule
    \end{tabular}%
    }
\end{table}

\paragraph{Baseline configuration.}
For Wan 2.1, we follow the baseline configurations used by SVG-EAR,
with the key SVG2 and SVG-EAR parameters summarized in
Table~\ref{tab:appendix-svg-config}. Their warm-up schedules follow
Table~\ref{tab:appendix-warmup}. 
For Sol-Attn, we use the diagonal estimator
with $\tau=0.52$, obtained by inverting the Gaussian approximation for a target
attention density of 30\%. We use the same Sol-Attn configuration for
HY3D-Internal. For a fair comparison, none of the methods applies special
treatment to condition tokens.

\begin{table}[!htbp]
    \centering
    \caption{SVG2 and SVG-EAR settings following the SVG-EAR evaluation setup.
    $Q_c$ and $K_c$ denote the numbers of query and key clusters.}
    \label{tab:appendix-svg-config}
    \small
    \setlength{\tabcolsep}{6pt}
    \renewcommand{\arraystretch}{1.12}
    \begin{tabular}{@{}lcccc@{}}
        \toprule
        & & & \multicolumn{2}{c}{TopP} \\
        \cmidrule(lr){4-5}
        Model & $Q_c$ & $K_c$ & SVG2 & SVG-EAR \\
        \midrule
        Wan2.1-14B-T2V & 300 & 1000 & 0.90 & 0.85 \\
        Wan2.1-14B-I2V    & 300 & 1000 & 0.90 & 0.85 \\
        \bottomrule
    \end{tabular}
\end{table}

\paragraph{Efficiency measurement.}
We report speedup relative to dense attention based on DiT denoising latency, including dense warm-up computation.
We do not report SpargeAttn's runtime because its implementation couples sparsification with low-precision attention, using INT8 or FP8 instructions for the QK and PV matrix multiplications, whereas the other methods use BF16, making the latencies not directly comparable. PISA's provided kernel cannot run efficiently on this model without modification. We therefore use the Sol-Attn kernel latency at the same attention density to estimate an upper bound on PISA's denoising speedup on HY3D-Internal.

\subsection{Implementation Details}
\label{app:implementation_details}

\paragraph{Operators and attention backend.}
Following the implementation practices of SVG2 and SVG-EAR, all Wan 2.1 and
HunyuanVideo runs, including the dense-attention baseline, use optimized operators
such as RoPE and normalization. Dense attention during warm-up uses
FlashAttention-3 (FA3). Minimax-H3-Base uses the operators provided by Diffusers, with FA3 for both the dense-attention baseline and dense warm-up steps.

\paragraph{Hardware and parallelism.}
Wan 2.1, HunyuanVideo, and HY3D-Internal run on 1 Hopper GPU, including the condition encoder and VAE where applicable.
Minimax-H3-Base runs on 8 Hopper GPUs: 1 handles condition encoding and VAE
decoding, while the remaining 7 execute the denoiser with Ulysses sequence
parallelism, which partitions the attention heads. 

\paragraph{Two-Pass exact selection kernel.}
\label{app:two-pass-selection}

FlashAttention uses online softmax and does not expose the full normalized
attention matrix needed for exact selection. We therefore use two passes:
the first returns the dense output and row-wise log-sum-exp values; the second
recomputes QK scores and uses these values to recover attention probabilities.
We fuse group-wise aggregation into the second pass and apply top-$K$ to the
aggregated attention mass, avoiding materialization of the dense attention
matrix.

\paragraph{Token-Sparse Attention kernel.}
\label{app:token-sparse-kernel}

We implement the token-sparse attention kernel in CuTeDSL with warp-specialized
producer and consumer warps. Producer warps resolve selected indices and gather
key and value rows asynchronously into shared memory. Consumer warpgroups apply
WGMMA operations and online softmax, while a multi-stage pipeline overlaps the
next gather with the current computation.

\clearpage
\section{More qualitative results}
\begin{figure}[H]
    \centering
    \includegraphics[width=\textwidth]{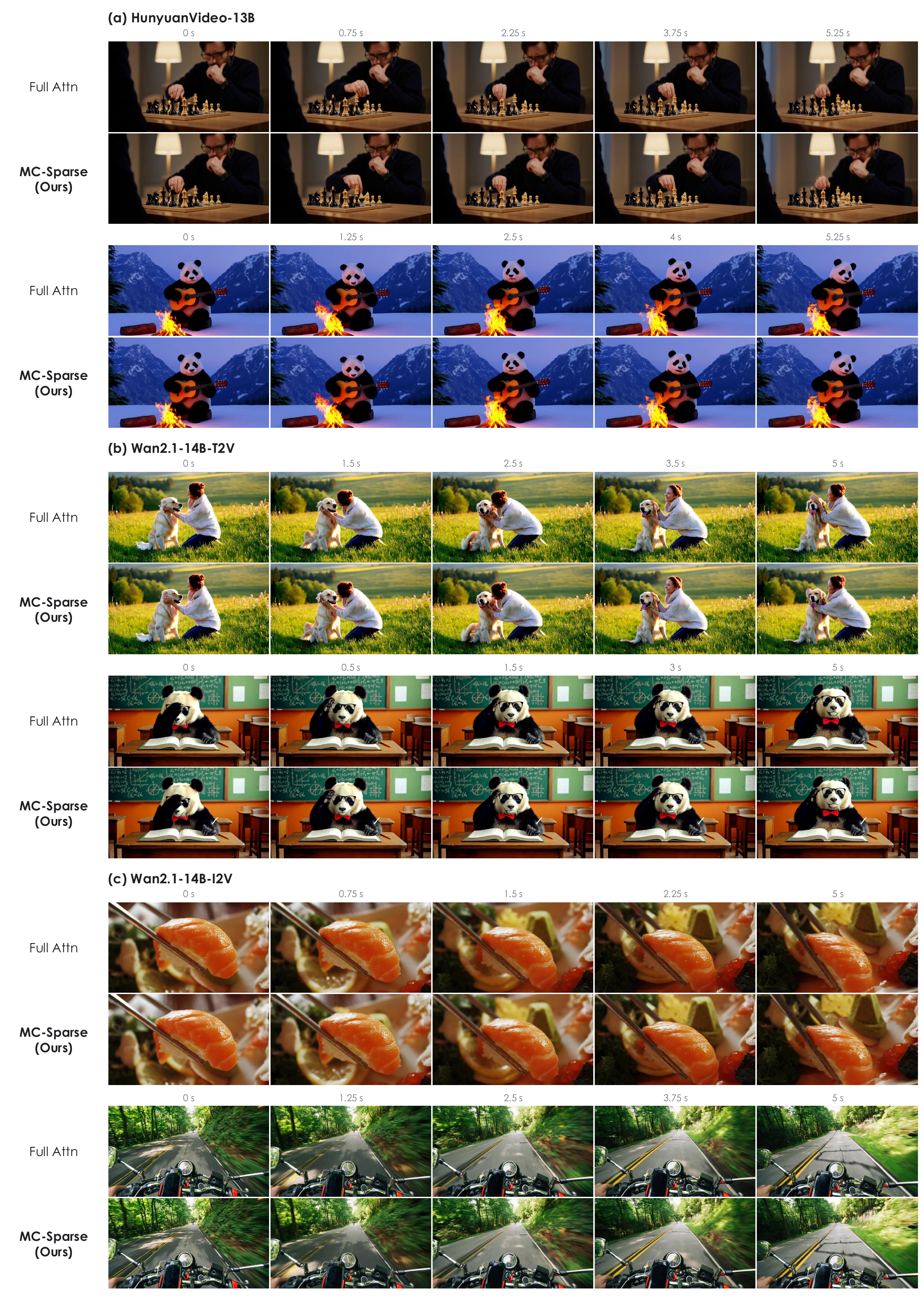}
    \caption{Qualitative comparison with full attention across video generation models.}
    \label{fig:supp_qualitative_comparison}
\end{figure}

\end{document}